\documentclass[letterpaper, 10 pt, journal, twoside]{ieeetran}

\usepackage{times}
\usepackage{epsfig}
\usepackage{graphicx}
\usepackage{amsmath}
\usepackage{amssymb}
\usepackage{color}
\usepackage{comment}
\usepackage{subfigure}
\usepackage{mathptmx}  
\usepackage{microtype}

\IEEEoverridecommandlockouts                              

\title{Nonrigid Optical Flow Ground Truth for Real-World Scenes with Time-Varying Shading Effects}

\author{Wenbin Li, Darren Cosker, Zhihan Lv and Matthew Brown

\thanks{\footnotesize Manuscript received February 28, 2016; Revised June 15, 2016; Accepted June 29, 2016. This paper was recommended for publication by Editor-in-Chief A. Bicchi and Editor J. Kosecka upon evaluation of the reviewers comments. This work was supported by the UK CAMERA EP/M023281/1 and EPSRC projects EP/K023578/1 and EP/K02339X/1. Our public evaluation website: https://vision.cs.bath.ac.uk/flow}

\thanks{\footnotesize Wenbin Li (the corresponding author) and Zhihan Lv with Department of Computer Science, University College London, UK.
        {\tt\small \{w.li, z.lu\}@cs.ucl.ac.uk}}%
\thanks{\footnotesize Darren Cosker with  Centre for the Analysis of Motion, Entertainment Research and Applications (CAMERA), University of Bath, UK.
        {\tt\small d.p.cosker@bath.ac.uk}}%
\thanks{\footnotesize Matthew Brown with Google, Mountain View, CA 94043, US.
        {\tt\small m.brown@bath.ac.uk}}%
\thanks{Digital Object Identifier (DOI): see top of this page.}
        \vspace{-4mm}
}

\begin{document}
\bstctlcite{IEEEexample:BSTcontrol}
\maketitle


\markboth{IEEE Robotics and Automation Letters. Preprint Version. Accepted June, 2016}
{Li \MakeLowercase{\textit{et al.}}: Nonrigid Optical Flow Ground Truth}

\begin{abstract}
In this paper we present a dense ground truth dataset of nonrigidly deforming real-world scenes. Our dataset contains both long and short video sequences, and enables the quantitatively evaluation for RGB based tracking and registration methods. To construct ground truth for the RGB sequences, we simultaneously capture Near-Infrared (NIR) image sequences where dense markers -- visible only in NIR -- represent ground truth positions. This allows for comparison with automatically tracked RGB positions and the formation of error metrics. Most previous datasets containing nonrigidly deforming sequences are based on synthetic data. Our capture protocol enables us to acquire real-world deforming objects with realistic photometric effects -- such as blur and illumination change -- as well as occlusion and complex deformations. A public evaluation website is constructed to allow for ranking of RGB image based optical flow and other dense tracking algorithms, with various statistical measures. Furthermore, we present an RGB-NIR multispectral optical flow model allowing for energy optimization by adoptively combining featured information from both the RGB and the complementary NIR channels. In our experiments we evaluate eight existing RGB based optical flow methods on our new dataset. We also evaluate our hybrid optical flow algorithm by comparing to two existing multispectral approaches, as well as varying our input channels across RGB, NIR and RGB-NIR.

\end{abstract}


\section{Introduction}

\IEEEPARstart{T}racking is a difficult task involved in many fields e.g. postproduction~\cite{Rotopp2016}, long term tracking~\cite{tang,APO,APO_JIFS}, reconstruction~\cite{godard2015multi} and interaction~\cite{tv}. The quantitative evaluation of optical flow algorithms is a difficult challenge -- particularly given long nonrigid scenes with natural noise. The \emph{Middlebury} benchmark~\cite{Middlebury} and the variations~\cite{moBlur,li2013nonrigid,vi_nc} are currently the most widely used \emph{Ground Truth} (GT) in the community. Tracking algorithms which use RGB/Color images may be submitted to the benchmark website for ranking and evaluation. However, this dataset is limited by the lack of object blur, complex nonrigid motion and long image sequences. Most of these limitations are due to the stop-motion method of capture: a scene is first captured under normal lighting; and then a second image of the same scene is captured using ultraviolet lighting. To address these limitations, Butler~\emph{et al.}~\cite{Sintel} proposed a dataset based on a 3D animated film \emph{Sintel}, which contains inter-frame GT through long sequences and geometric blur under different renderings. However their inherent limitation is the use of synthetic sequences, which lacks real-world photometric effects and textural properties. Similar to \emph{Sintel}, Garg~\emph{et al.}~\cite{Garg} rendered synthetic video sequences accompanying with GT by projecting the scene motion (\emph{Motion Capture}) of a realistic waving flag onto an image plane. Although KITTI~\cite{Geiger} benchmark enables the evaluation in real-world street scenes, there is still a lack of nonrigid GT for long sequences.

\begin{figure}[t!]
\begin{center}
\includegraphics[width=0.99\linewidth]{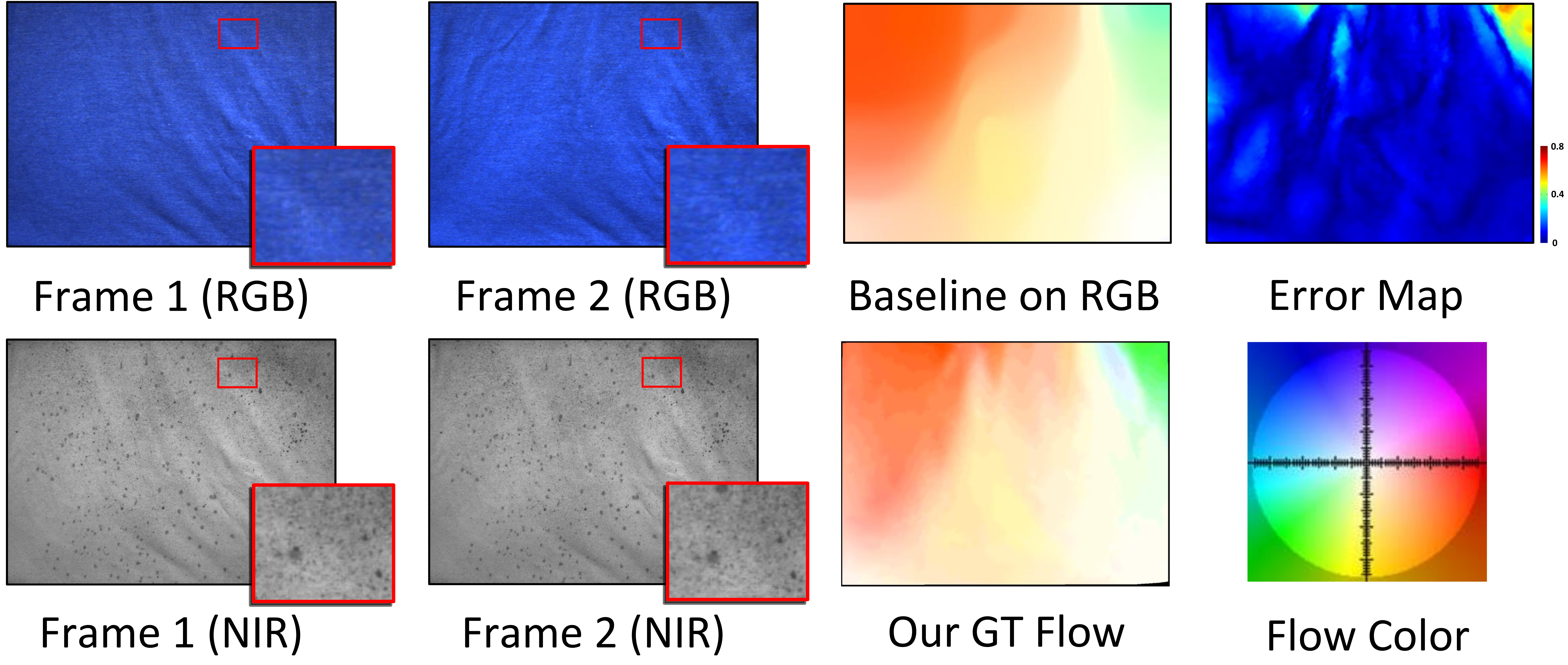}
\end{center}
\caption{\small A baseline algorithm~\cite{LDOF} is performed on the RGB channel of one of our ground truth sequences \emph{featureless}. The figure highlights a dense NIR GT patch -- used to evaluate RGB based tracking -- in an otherwise near-textureless RGB region. \vspace{-4mm}}
\label{fig:introImg}
\end{figure}

In this paper, we propose such a GT dataset -- allowing for the first time the benchmark of dense tracking algorithms on real-world nonrigidly deforming scenes captured at video rate. Sequences may be tracked using the RGB channel, and their performance measured against the GT. The key insight to capture such a dataset is the use of multispectral imaging -- in particular, RGB\&Near-Infrared (RGB-NIR) imaging which has recently been shown useful in computer vision, e.g. multispectral SIFT ~\cite{Matt}, image dehazing~\cite{dehazing} and registration~\cite{registration,moblur_nc}.  A property of such imaging is the ability to apply markers visible in one spectrum (e.g. NIR), but invisible in another (e.g. RGB). Therefore, an algorithm can be applied to the RGB sequence alone, and its performance is then compared to the invisible markers in the NIR channel. To accompany with the data, we provide an evaluation platform which allows users to download the RGB data, upload their tracking results and then view the accuracy versus other methods on our GT.

The second focus of our paper is to investigate how multispectral (RGB-NIR) imaging might improve the quality of tracking, by proposing a multispectral optical flow formulation. The variational optical flow model began with the pioneering work of Horn and Schunck~\cite{HS} and Lucas and Kanade~\cite{LK}. Some complementary concepts have since been developed to deal with the shortcomings of their original models such as spatial discontinuities~\cite{BA}, large displacements~\cite{LDOF}, motion detail loss through coarse-to-fine minimization~\cite{MDP} and local smoothness~\cite{LME}. Of these methods, Xu~\emph{et al.}'s (MDP)~\cite{MDP} approach is currently amongst top 3 (by average) in the \emph{Middlebury} evaluation while the Li~\emph{et al.}'s (LME)~\cite{LME} approach has state-of-the-art performance given nonrigid surface motion~\cite{Garg}. However, all of these methods are applied on image pairs within the visible spectrum (RGB/Color) and are sensitive to motions in large featureless regions in which the basic \emph{Intensity Consistency} assumption is weakened.

To take an advantage of extra spectrums, Markandey and Flinchbaugh~\cite{msof} consider the IR image within an optical flow method, which solves a system of two data terms (RGB/Grayscale and IR). They assume an equal contribution (a known fixed weight) from both channels. This may reduce the precision in some cases (Fig.~\ref{fig:weights}). Barron and Klette~\cite{cof} propose an approach using all three individual color channels, and shows improvement over the grayscale. 


\subsection*{Contributions}

\begin{figure}[t!]
\begin{center}
\includegraphics[width=0.96\linewidth]{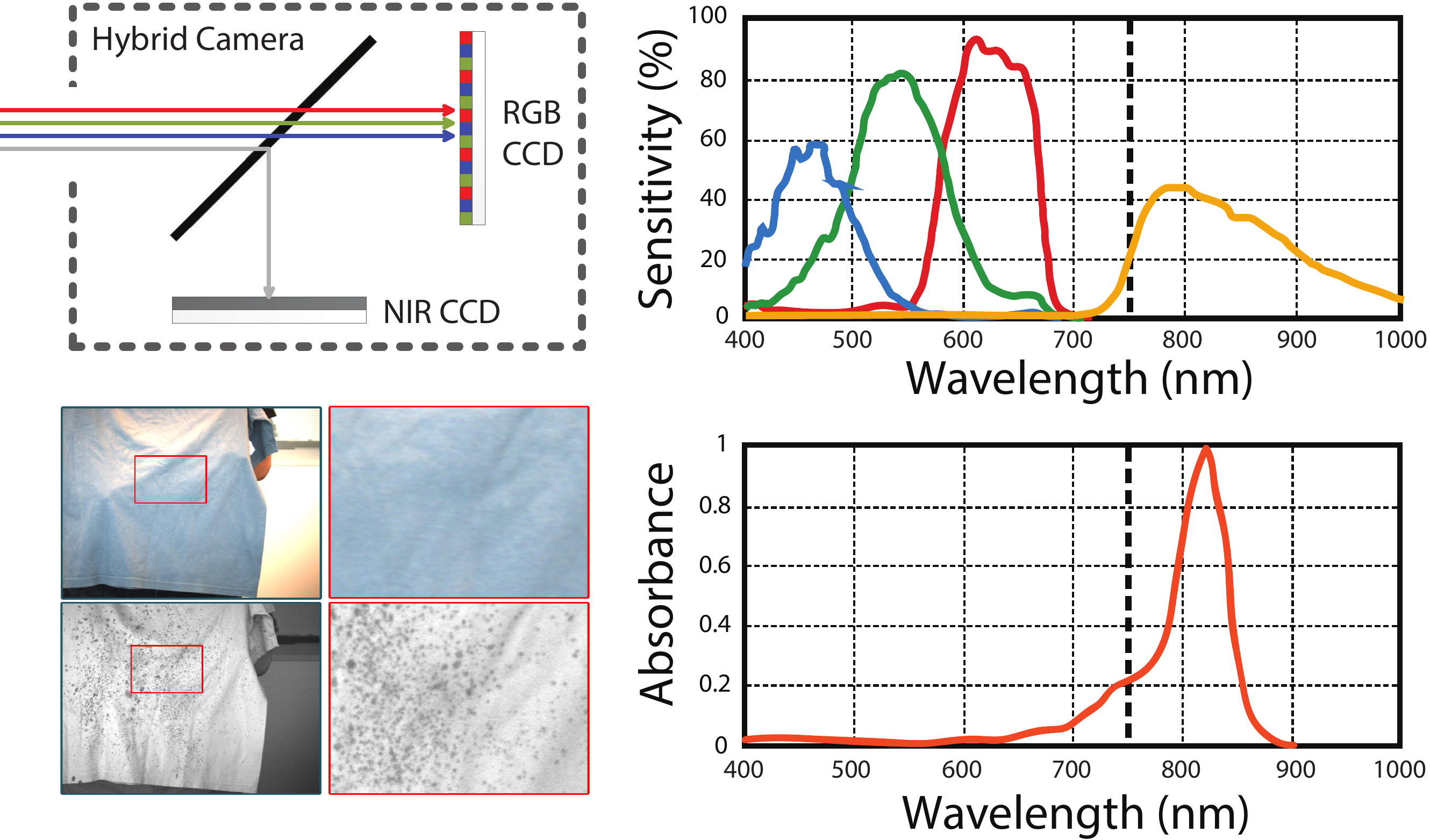}
\end{center}
\caption{RGB-NIR Camera and the NIR visible dyes. \textbf{Top Left}: The inside structure of the camera. \textbf{Bottom Left}: Sample images captured by the RGB CCD sensor and NIR CCD sensor respectively. \textbf{Top Right}: The relative transmittance of our RGB CCD sensor and NIR CCD sensor (yellow). \textbf{Bottom Right}: The absorbance of the NIR visible dyes respect to various wavelength.\vspace{-4mm}}
\label{fig:imaging}
\end{figure}

To summarize there are two major contributions in our paper: \emph{\textbf{(1)}} we present a nonrigid GT dataset (Fig.~\ref{fig:introImg}) for RGB image based dense tracking (e.g. optical flow) methods, and an evaluation website allowing users to rank the performance of their method versus others. The dataset contains dense inter-frame correspondences from eight short and five long sequences with varying photometric properties; and \emph{\textbf{(2)}} we present a multispectral (RGB-NIR) optical flow model (\emph{vnflow}). Within this model, we propose a novel weighting scheme which adoptively selects the best available image features in either the RGB or the complementary NIR channel to enhance motion analysis.

In our experiments, we evaluate ten existing RGB based optical flow methods on our dataset - ranking them based on various statistics (the same presented on our evaluation website). We then turn the attention to our \emph{vnflow} method which illustrates the potential benefit of using combined spectra (e.g. RGB-NIR) for optical flow estimation.



\section{Non-Rigid Ground Truth Dataset}
\label{sec:dataset}

\emph{Ground Truth} (GT) for RGB/Color optical flow is difficult to capture -- how does one simultaneously acquire an invisible set of GT positions upon which to evaluate algorithm performance on the visible RGB channel? One important advance in this area was proposed by Baker~\emph{et al.} with the introduction of the \emph{Middlebury} benchmark~\cite{Middlebury}. Due to their contribution, the optical flow community has rapidly developed in recent years. However, Baker~\emph{et al.} also point out limitations of their work -- the lack of object blur and occluded motion -- which are discussed in more recent state-of-the-art datasets~\cite{Sintel}. The main limitations of current benchmarks , which we address in this work, are as follows:


\paragraph*{\textbf{Long Image Sequences}} As discussed in~\cite{Sintel}, most of the \emph{Middlebury} sequences are short in length, which leads to a lack of evaluation on long term correspondence. While \emph{Sintel} provides long synthetic sequences (more than 50 frames) and GT for each pair of frames, our dataset provides long sequences from real-world objects -- thus exhibiting realistic photometric effects and textural properties.


\paragraph* {\textbf{Realistic Noise}} The lack of realistic blur is a common issue in both \emph{Middlebury} and \emph{KITTI}. Our dataset includes realistic camera blur and other noise, e.g. strong shadows, reflectance and illumination changes. 


\paragraph* {\textbf{Complex Nonrigid Motions}} Unlike \emph{Middlebury} and \emph{Sintel}, our dataset is specifically focused on nonrigid motion, containing examples of stretching, large bends and creases.


\subsection{RGB-NIR Imaging System}

In order to acquire our GT, we construct a controllable scene (i.e. lighting and motion properties) using an \emph{RGB-NIR Imaging System} and \emph{NIR Visible Dyes}.


\paragraph* {\textbf{RGB-NIR Camera}} In this paper, a hybrid camera (JAI AD-080GE) is used to capture both RGB and NIR images from the same scene simultaneously. Fig.~\ref{fig:imaging} shows internal construction of the camera, where natural light is split onto the RGB and NIR CCD sensors respectively. As opposed to experimental bench-based RGB-NIR beam-splitter setups~\cite{Cao}, the overall system is both compact and portable. Such a system simultaneously captures a series of continuous images in both the RGB and NIR channels at 20 FPS.

\begin{figure}[t!]
\begin{center}
\includegraphics[width=0.97\linewidth]{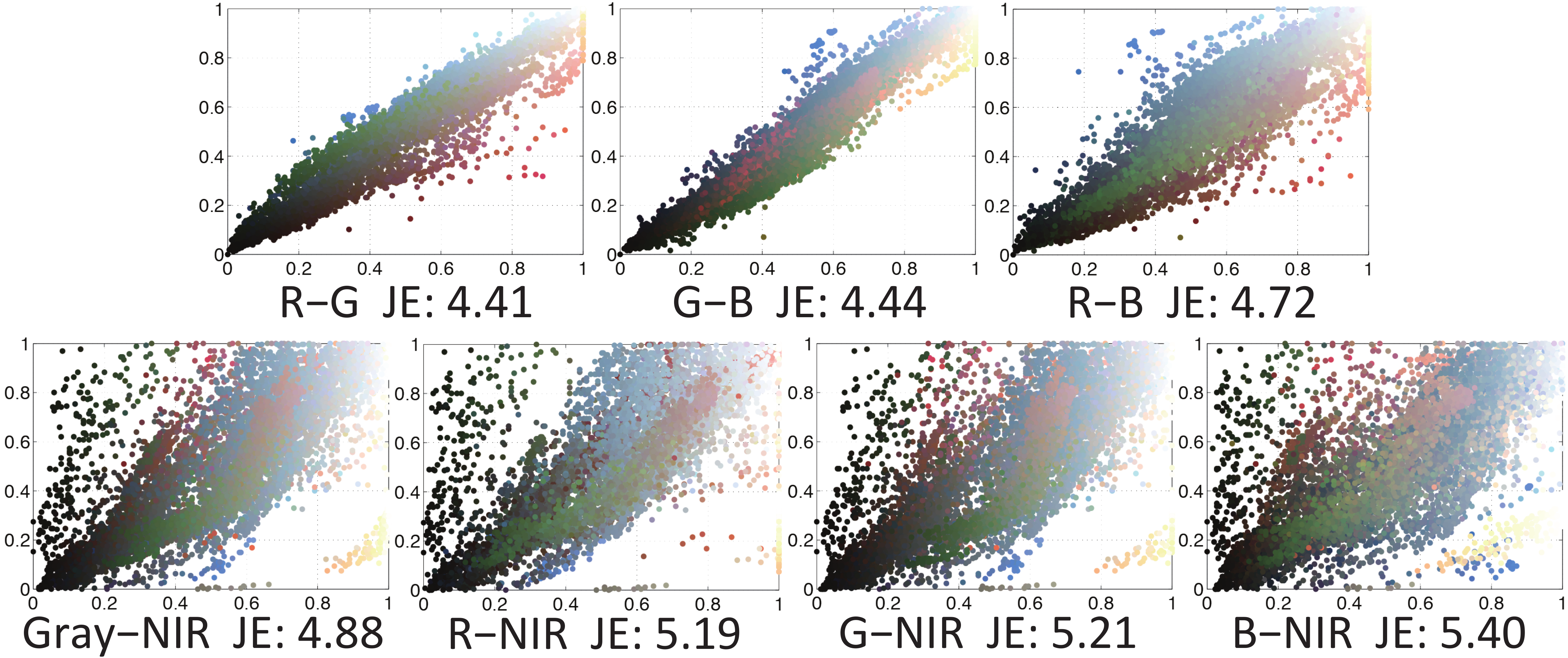}
\end{center}
\caption{Pairwise distributions for the RGB and NIR channels of 20,000 sampled patches from our ground truth dataset. \vspace{-4mm}}
\label{fig:pairContr}
\end{figure}

\paragraph* {\textbf{NIR Visible Dyes}} In order to generate dense features on object surfaces for our GT dataset, we utilize \emph{NIR Visible Dyes} (NIR819D, QCR Solutions Corp.) which absorb the spectrum in a range of approximately 700 to 870 nm with a peak at around 819 nm. Our \emph{NIR Visible Dyes} are spread onto object surfaces in order to generate fine patterns of which the diameter is within 1 mm, with a maximum 2 mm distance between any pair of neighboring patterns. Fig.~\ref{fig:imaging} shows dense patches painted by our dyes that are visible in the NIR channel while remaining invisible in the RGB channel. To illustrate the statistical dependencies of the patches between different bands, $20,000$ RGB-NIR patches ($3 \times 3$ pix.) with the dyes applied are randomly selected and plotted as pairwise distributions using joint entropy in Fig.~\ref{fig:pairContr}. Note that we compute the joint entropy as $H(X,Y)=-\sum_{X,Y}P(X,Y)log_2{[P(X,Y)]}$. It is observed that the joint entropy of \{R,G,B,Gray\}-NIR is larger than between the visible bands (R,G,B). Therefore, the \emph{NIR Visible Dyes} provide richer visible information apart from RGB channel -- making it suitable for a GT basis especially in largely textureless RGB channel regions. The Middleburry benchmark adopts UV-flourescent dye, only visible under the UV lighting. This leads to an issue that they have to stop the object movement and switch to UV lighting when they need to capture the dye-featured image. In this case, their sequences may lose the blurry photometric effects. However, our \emph{NIR Visible Dyes} together with \emph{RGB-NIR Camera} allows a continuous capture (up to 20 FPS) which is able to preserve the blur and other real-world photometric effects possible. In practice, our dyes give the best invisibility on the cotton surface, but are hard to remain completely invisible on other materials. Our dyes cannot be applied to human for long because it would harm the skin in some cases.

\paragraph* {\textbf{Motion Control Component}} To precisely control the displacement of objects in our GT scenes, a motion control mechanism is constructed using LEGO NXT Mindstorms robotics kits which produce controllable and uniform inter-frame movements for our GT surfaces.

In the following section, we describe dataset construction together with our proposed evaluation methods.

\begin{figure*}[t!]
\centerline{
\includegraphics[width=0.97\linewidth]{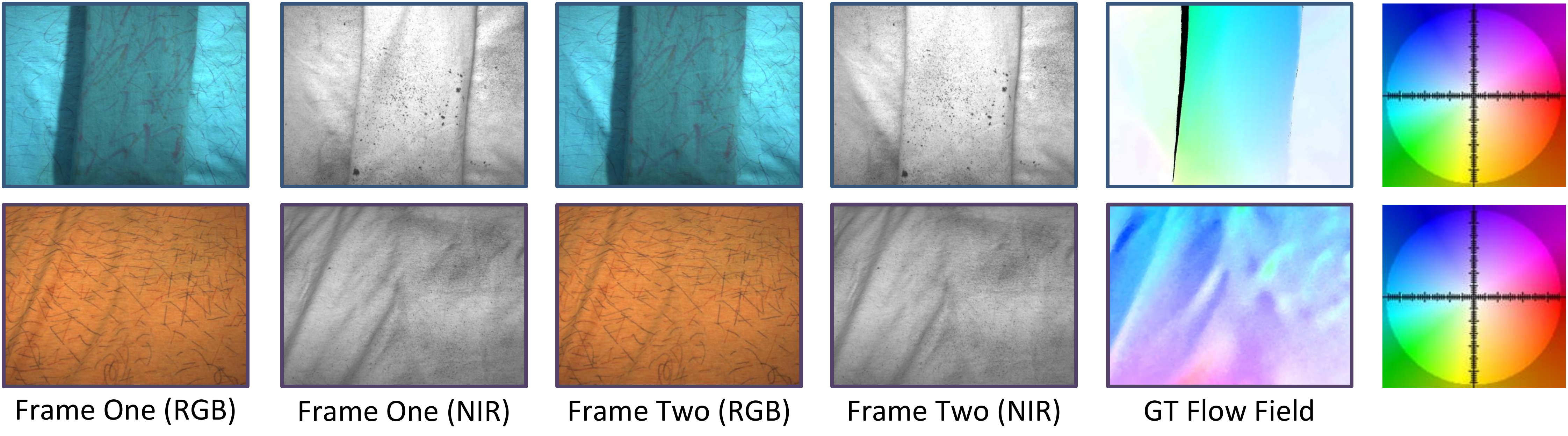}
}
\caption{Examples from our ground truth dataset. \textbf{Top Row:} \emph{str.shadow} contains strong shadows and subtle nonrigid motion. \textbf{Bottom Row:} \emph{crush} (frames 38 and 39) is a video sequence containing complex nonrigid deformations and self occlusions.\vspace{-6mm}}
\label{fig:samples}
\end{figure*}


\subsection{Ground Truth Estimation}

Once we have obtained the corresponding pairs of RGB and NIR images, we use the feature-rich NIR channel to construct a dense GT flow field. In this subsection we describe this process and other important properties in detail.


\paragraph* {\textbf{Image Properties}} Our \emph{RGB-NIR camera} captures images at $1296\times966$ pixels. The \emph{Motion Control Component} of our system allows us to precisely range motions from subpixel to 40 pixels. Similar to \emph{Middlebury}, all the captured RGB sequences are downsampled by a factor of 3, resulting in an image size of $432\times322$ after the \emph{Subpixel Motion Estimation} step (presented later in this subsection).


\paragraph* {\textbf{Data Acquisition}} To capture the data properly, we set up a capture system using our \emph{Motion Control Component} and \emph{RGB-NIR Camera}. The motor (NXTMotor) of the component is able to precisely rotate by 1 degree step. Together with the Lego bricks and bars, \emph{Motion Control Component} precisely controls the motion of the object surfaces in the scale of $[1, 46]$ cm. Most of the motion represented in the dataset is parallel to the camera plane. Furthermore, Our camera is distortion free and follows a pinhole model. In this case, the calibration aims to find out the relation $f$ between the object movement $M_{MCC}$ (in cm) and the pixel displacement $M_{p}$ within the image space. Here we have $M_{p} = f(M_{MCC})$. In practice, we fix the distance (1.5m) between the objects and camera while the camera forward direction is perpendicular to the object surface. We then capture the surface motion with a certain $M_{MCC}$; and manually measure the $M_{p}$ from the image. We repeat this process before capturing each of sequences and obtain the size of search window $2M_{p}\times2M_{p}$ pix. for the following \emph{Pixel Correspondence} estimation.

\paragraph* {\textbf{Pixel Correspondence}} We use a parfum spray to generate fine patterns on the objects. In most cases, the diameters of such patterns are smaller than 1 mm, corresponding to approximately 1 pixel of the image (Fig.~\ref{fig:weights} (Left)). And those patterns are still highly variable in terms of intensity and shape. Therefore pixel correspondences are achieved by matching the dye patterns between neighboring NIR images. Unlike the Color-\emph{SSD} tracker used in \emph{Middlebury}, we consider both intensity and shape. A SIFT descriptor with 128 dimensions is computed for each pixel in an image. We nominate a GT match between pixels where the \emph{Euclidean Distance} of their SIFT vectors is smallest within a given search window. This window size ($2M_{p}\times2M_{p}$ pix.) is predefined using the maximum motion in the \emph{Motion Control Component}. To improve robustness we examine the matched results across adjacent frames. A correspondence is labeled with a value ``\emph{NAN}''  (Not-A-Number) if the intensity difference between the forward matched result and the backward matched result is greater than a threshold. The region mask containing ``\emph{NAN}'' values is recorded as an occlusion map. Note that we do not apply an existing optical flow method onto the NIR images to estimate the correspondences. Although the optical flow is able to give us the per-pixel dense correspondence, the encoded smoothness term may overly smooth the motion at the object boundaries and small motion details. This would further reduce the precision of the GT.


\paragraph* {\textbf{Subpixel Motion Estimation}} After obtaining GT pixel correspondence, we follow the \emph{Middlebury} subpixel motion estimation process. We apply the \emph{Lucas-Kanade} kernel~\cite{LK} to each search window for subpixel motion using 1/20 pixel precision. We then calculate the average of up to 9 motion vectors in each $3\times3$ window in order to down-sample the motion field to dimension $432\times322$.


\paragraph* {\textbf{Realistic Noise}} The controllable nature of our \emph{RGB-NIR Imaging System} allows us to incorporate varieties of noise and artefacts into our GT dataset. We increase the exposure time of the RGB CCD sensor to bring object blur into the visible channel, while using a suitably fast exposure time on the NIR CCD sensor to capture a corresponding blur-free image. Alternatively, defocus blur could also be obtained by modifying the aperture settings. Shadow and illumination changes are generated using infrared-free light (LED lighting), leading to realistic shadows in the RGB channel without affecting illumination in NIR channel (Fig.~\ref{fig:samples}).


\paragraph* {\textbf{Sequence Descriptions}} Here we provide labels for each of the sequences in our dataset, as well as brief descriptions of their characteristics. Our dataset contains two types of GT sequence -- \emph{Short Sequences} and \emph{Long Sequences}. We capture eight short sequences in total, each of which contains ten frames with dense GT between middle pair of images. Each sequence is captured so as to include specific common image properties. In terms of sequence naming, \emph{single} contains nonrigid motion of single object. \emph{illumination} contains strong reflectance and illumination change while both \emph{mObjs} and \emph{triObjs} contain multiple objects with nonrigid movement. \emph{featureless} contains small motions for a featureless object surface while \emph{crease} contains a large crease on multiple objects. \emph{blur} and \emph{str.shadow} contain both camera blur and strong shadows respectively. In addition, five longer sequences are captured with dense inter-frame GT for every neighboring image pair. Each sequence contains 50 frames and includes multiple realistic photometric effects and nonrigid motion. \emph{mBlur} contains focus blur, motion blur and large displacements, while \emph{circle} contains complex creases. \emph{crush} presents a large crushing movement with self occlusions and \emph{stretch} shows elastic deformation. Finally, \emph{wave} presents a real-world waving cloth. We also provide training set which contains 3 short and 3 other long GT sequences.

Fig.~\ref{fig:samples} shows two sample sequences (\emph{str.shadow} and \emph{crush}) from our dataset where tracking algorithms are executed on the RGB data, and the NIR channel - with the aid of NIR visible dyes - provides our GT flow fields upon which to compare to the RGB flow fields. In the following section, we introduce our evaluation methods along with the public website to openly evaluate algorithms.


\subsection{Evaluation Methods and Statistics}
\label{sec:statistic}

Similar to \emph{Middlebury}, we provide tests of \emph{Endpoint Error} (EE) and \emph{Angle Error} (AE). Users are expected to download the RGB data from our evaluation platform, and compute flow fields between all frames in the \emph{Long Sequences} and for one image pair for each sequence in the \emph{Short Sequences}. Users then upload their result and our evaluation system compares it to the GT flow fields calculated on our NIR channel (which includes the NIR visible dyes). For robustness statistics, we perform \emph{Average} (Avg.), \emph{Accumulated} (Acc.), \emph{Standard Deviations} (SD), \emph{RX} and \emph{AX}~\cite{Middlebury} where \emph{RX} presents the percentage of pixels that have an error reading above X; And \emph{AX} is for the accuracy of the error reading at the Xth percentile, after sorting the errors from low to high. Avg., SD and \{A50, A75, A99, A100\} are given for both EE and AE; \{R0.5, R0.75, R1, R2\} are performed for EE; Acc. is calculated for EE in long sequences only; \{R2, R5, R7.5, R10\} are computed for AE.

As shown in Fig.~\ref{fig:sceneShotShort}, we generate a comparison table for cross-evaluation of user uploaded flow field results against any other methods previously uploaded to our evaluation system. For long sequences, we can plot results selected by the user with respect to a specific frame index.


\section{RGB-NIR Variational Optical Flow Model}

In the previous sections, we described a GT dataset and evaluation website for algorithms operating on RGB data. In this section, we now slightly change focus and introduce a novel algorithm which combines both RGB and NIR channels in such a way as to maximize the distinguishing information from each channel. Certain visual information can be poorly represented in an RGB channel. It is therefore prudent in many cases to also consider the NIR channel (and vice-versa). In our evaluation section we examine these properties in more detail. Note that for fairness, our public RGB-only evaluation website does not include results of our \emph{vnflow} or any future multispectral methods.

Our algorithm considers a pair of consecutive frames in an image sequence. The current frame is denoted by $I_{1}(\textbf{x})$ and its successor is $I_{2}(\textbf{x})$ where $I=(V,N)^T$, $\{V:\Omega \subset \mathbb{R}^3 \to \mathbb{R}\}$ represents a rectangular image in the RGB channel and $\{N:\Omega \subset \mathbb{R}\}$ denotes a rectangular image in the NIR channel. Both $V$ and $N$ are aligned and share the same Cartesian coordinate where $\textbf{x} = (x,y)^{T}$ is a pixel location. The optical flow displacement between $I_{1}(\textbf{x})$ and $I_{2}(\textbf{x})$ is defined as $\textbf{w} = (u,v)^{T}$. Our proposed optical flow approach leads to the following energy function:

\begin{eqnarray}
E(\textbf{w}) = (1-\lambda(\textbf{x}))E_{V}(\textbf{w}) + \lambda(\textbf{x}) E_{N}(\textbf{w}) + \gamma E_{S}(\textbf{w})
\label{eq:energy}
\end{eqnarray}

where the \emph{Visible RGB Energy} $E_{V}(\textbf{w})$ contains both \emph{Intensity Constancy} and \emph{Gradient Constancy} assumptions between the visible components $V_{1}(\textbf{x})$ and $V_{2}(\textbf{x})$ of the images while our main contribution i.e. \emph{Invisible NIR Energy} is represented as the term $E_{N}(\textbf{w})$. A high-order regularization $E_{S}(\textbf{w})$ is adopted.


\paragraph* {\textbf{Visible RGB Energy}} Following the \emph{Intensity Constancy} assumption, we assume that the intensity of a pixel is not varied by its displacement throughout an image sequence. In addition, we also make a \emph{Gradient Constancy} assumption~\cite{LDOF} to provide additional stability where pixel intensity is violated by illumination changes. The \emph{Visible RGB Energy} term encoding these assumptions is thus formulated as:

\begin{align}
E_{V}(\textbf{w}) &= \int_{\Omega}  \phi ( \left \| V_{2}(\textbf{x}+\textbf{w}) - V_{1}(\textbf{x})\right \|^{2})d\textbf{x} \nonumber\\
&+\theta \int_{\Omega}  \phi ( \left \| \nabla V_{2}(\textbf{x}+\textbf{w}) - \nabla V_{1}(\textbf{x})\right \|^{2})d\textbf{x}
\end{align}


For robustness against occlusions and boundary blur, we apply the increasing concave function $\phi(s^{2}) = \sqrt{s^{2} + \epsilon^{2}}$ with $\epsilon = 0.001$ to solve this formation. The remaining term $\nabla = (\nabla_{x},\nabla_{y})^{T}$ is a spatial gradient and $\theta\in[0,1]$ denotes a linear weight. The smoothness term is a dense pixel based regularizer that penalizes global variation. The objective is to produce a globally smooth constraint:

\begin{eqnarray}
E_{S}(\textbf{w}) = \int_{\Omega} \phi(\left \| \nabla u \right \|^{2} + \left \| \nabla v \right \|^{2})d\textbf{x}
\end{eqnarray}


\paragraph* {\textbf{Invisible NIR Energy}} A \emph{visible RGB Energy} term is widely used in optical flow~\cite{LDOF} but error-prone in featureless regions or unclear boundaries. We therefore propose to inspect additional spectral channels given these situations. We include an \emph{Invisible NIR Energy} term as a complementary assumption to the classic framework, namely to introduce additional texture information to  optical flow estimation. Similar to the RGB \emph{Intensity Constancy}, we assume that the pixel intensity in the NIR channel is not changed by displacement, which yields an energy term as follows:

\begin{eqnarray}
E_{N}(\textbf{w}) = \int_{\Omega}  \phi ( \left \| N_{2}(\textbf{x}+\textbf{w}) - N_{1}(\textbf{x})\right \|^{2})d\textbf{x}
\end{eqnarray}

Where the term $E_{N}(\textbf{w})$ presents the continuous energy in the NIR channel. Note that both terms $E_{V}(\textbf{w})$ and $E_{N}(\textbf{w})$ share the same spatial smoothness regularizer. 
\begin{figure}[t!]
\centerline{
\includegraphics[width=0.95\linewidth]{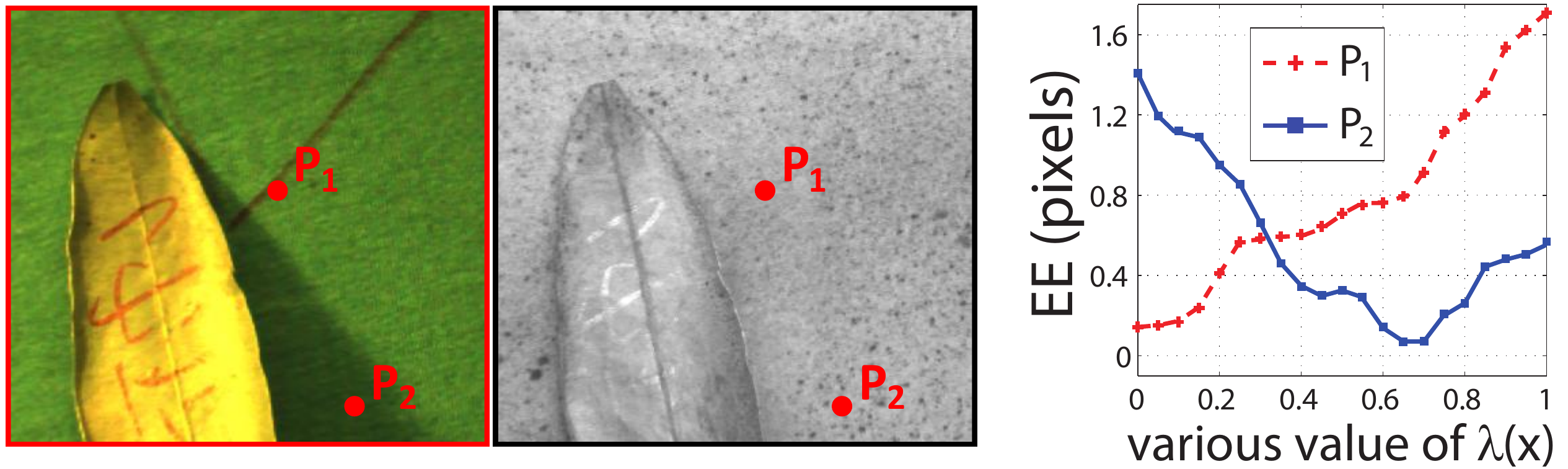}
}
\caption{\emph{Endpoint Error} (EE) affected by varying weight $\lambda(\textbf{x})$. \textbf{Left}: A patch of \emph{LeafShadow} is shown where two points $P_1$ and $P_2$ are plotted in RGB and NIR channels respectively. \textbf{Right}: EE for both points $P_1$ and $P_2$ are plotted by varying weight $\lambda(\textbf{x})$.}
\label{fig:weights}
\end{figure}


\paragraph* {\textbf{Detail-Aware Weight $\lambda(\textbf{x})$}} In Fig.~\ref{fig:weights} (Left) we show an image patch in which two points $P_1$ and $P_2$ are plotted. The small region centered on $P_2$ contains soft shadow in the RGB channel but has more distinguishing features in the NIR channel. For the point $P_1$, the situation is opposite. The \emph{Endpoint Error} (EE) with respect to the different $\lambda(\textbf{x})$ values are plotted in Fig.~\ref{fig:weights} (Right). We observe that plain texture leads to larger errors in the optical flow estimation. Dynamically taking more contribution from the channel containing more detailed texture is therefore adopted in our method.

\subsection{Minimization Framework}

Prior to energy minimization, \emph{$\lambda(\textbf{x})$ Initialization} is performed to improve overall optical flow energy in featureless regions. A numerical scheme is then applied to minimize the continuous RGB-NIR energy within a pyramidal framework. Both steps are described in following sections.


\paragraph* {$\lambda(\textbf{x})$ Initialization.} Inspired by the kernel-based edge detector where an \emph{Intensity Gradient} is used to represent geometric information in the texture space, we define a weight $\{\lambda(\textbf{x}):\mathbb{R} \mapsto [0,1]\}$ using an \emph{Intensity Gradient} as follows:

\begin{eqnarray*}
\lambda(\textbf{x}) = \left ( 1+ \exp \left \{-a \left ( \frac{\left |\nabla N_1(\textbf{x})\right |}{\left |\nabla V_1(\textbf{x})\right |+\left | \nabla N_1(\textbf{x})\right |}-b \right) \right \}\right )^{-1}
\end{eqnarray*}

where $\textbf{x}$ denotes a pixel location while $\nabla = (\nabla_{x},\nabla_{y})^{T}$ presents the intensity gradient  calculated using a $3 \times 3$ \emph{Sobel Kernel}; $a$ and $b$ are parameters chosen to be 10 and 0.5 respectively. The weight $\lambda(\textbf{x})$ is intensity-dependent and can be precalculated before energy minimization. Given an $n$-level image pyramid, the input images $I_1$, $I_2$ and the weight map $\lambda(\textbf{x})$ are resized to the same scale on each level. These are denoted by $I_1^i=(V_1^i,N_1^i)^T$, $I_2^i=(V_2^i,N_2^i)^T$ and $\lambda^i$, and are used in the following energy minimization phase.


\paragraph* {\noindent RGB-NIR energy optimization.} In this process, we aim to find the global minimum of the energy in Eq.~(\ref{eq:energy}) which is continuous but highly nonlinear. We need to remove the nonlinearity and obtain the final linear system. Thus, we apply nested fixed point iterations on $\textbf{w}$ by mainly following the numerical scheme in~\cite{Brox}. In the implementation, the image pyramid is constructed using a downsampling of 0.75. The final linear system is solved with successive over-relaxation. For more details of our optimization scheme, please refer to the supplementary document.

\section{Experiments}

In this section, we evaluate \textbf{\emph{(1)}} eight publicly available optical flow algorithms from \emph{Middlebury} using our nonrigid GT dataset (executed on the RGB channel, and compared against the NIR GT flow fields), and \textbf{\emph{(2)}} our proposed multispectral optical flow method (\emph{vnflow}) comparing to two multispectral approaches, highlighting the potential benefits of \emph{Detail-Aware Weighting}.

\begin{figure*}[t!]
\centerline{
\includegraphics[width=0.97\linewidth]{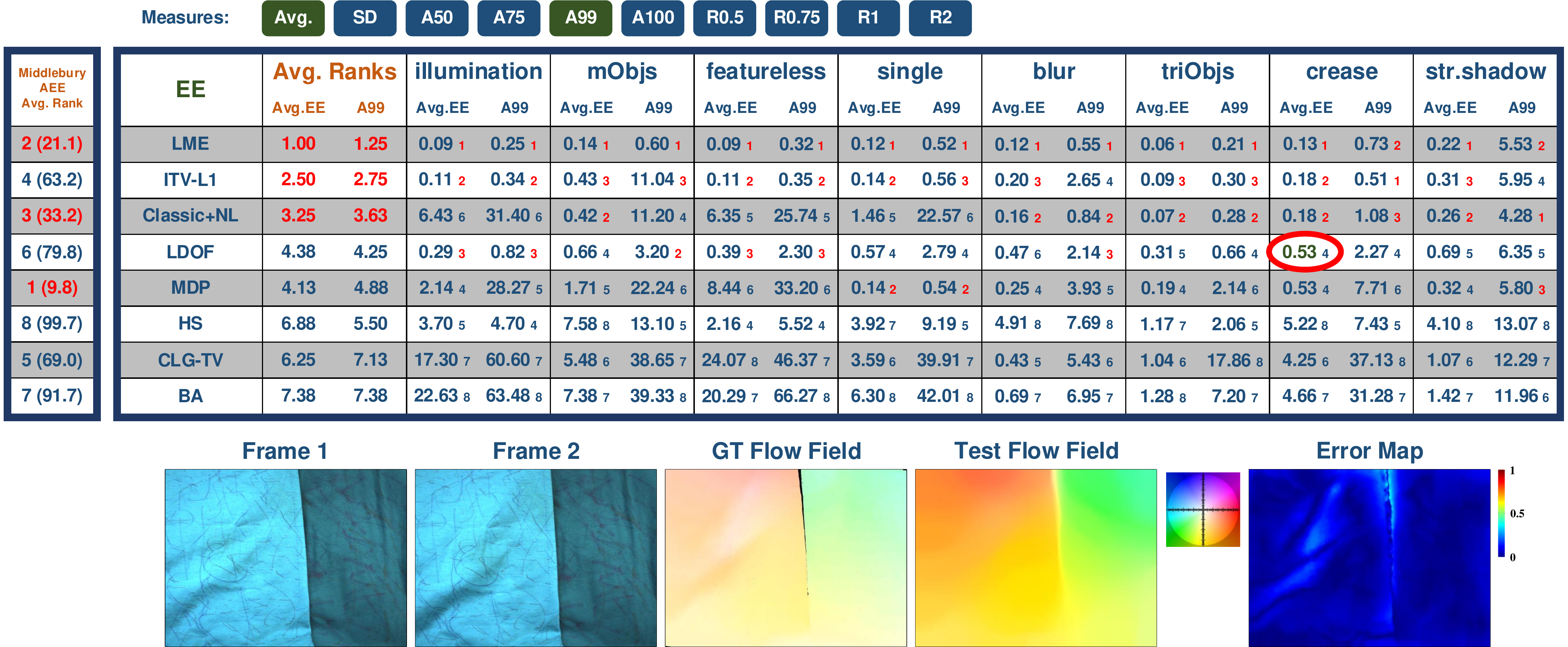}
}
\caption{Screen shot of our public evaluation website for RGB/Color based dense tracking/optical flow algorithms. Sample scores for the \emph{short} sequences are shown, and we demonstrate the \emph{Endpoint Error} (EE) evaluation. Multiple statistics/measures (Sec.~\ref{sec:statistic}) can be manually selected on the top of the table and illustrated as sub-columns within a sequence where the subscripts show the rank in that sub-column. The user can mouse-click any of the results to show sequence details, the proposed flow field and the error map against the ground truth (as shown on the bottom row). All methods are listed in order of their average rank (Avg. Ranks).\vspace{-3mm}}

\label{fig:sceneShotShort}
\end{figure*}

\begin{figure*}[t!]
    \centerline{
    \subfigure[\textbf{Table View} shows quantitative evaluation on all long sequences. The user can mouse-click any result to bring up the details (\textbf{Right Graph}), in which they are plotted w.r.t. the frame index. Any node within the graph can be clicked to show the visual comparison (ground truth, the proposed flow field and error map) for a specific frame index. ]{\includegraphics[width=0.97\linewidth]{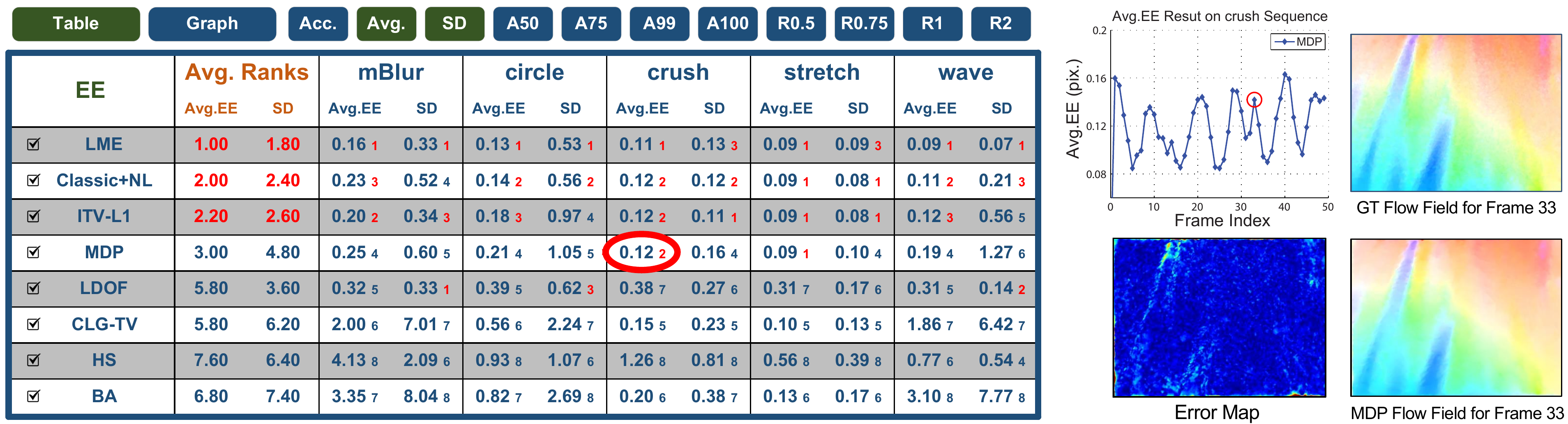}}
    }
    \centerline{
    \subfigure[\textbf{Graph View} plots details for each sequence. The user can select multiple baseline methods by clicking their checkboxes then clicking the \emph{Graph} option on top of the table. The measure details, e.g. Avg.EE and Acc.EE, are plotted onto the downloadable graphs for each sequence.]{\includegraphics[width=0.97\linewidth]{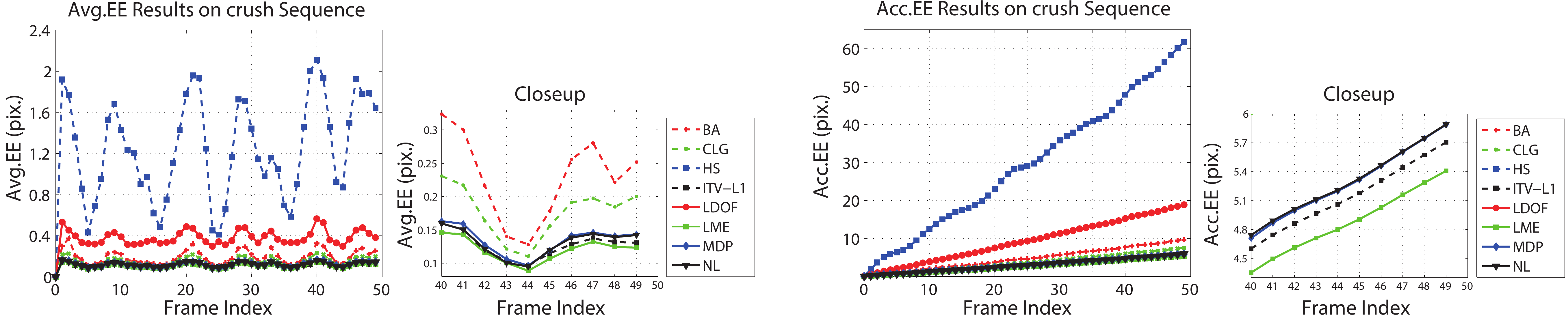}}
    }
    \caption{Screen shot of our public evaluation website for long sequences, illustrating the \emph{Endpoint Error} (EE) evaluation.\vspace{-2mm}}
    \label{fig:sceneShotLong}
\end{figure*}

\begin{figure*}[t!]
\centerline{
\includegraphics[width=0.98\linewidth]{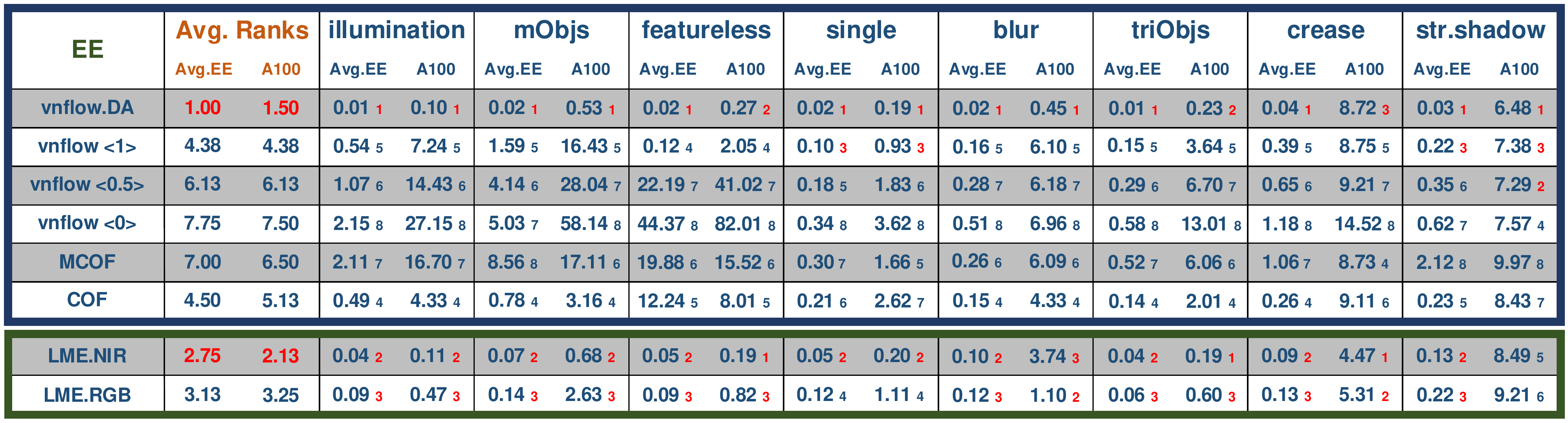}\vspace{-1mm}
}
\caption{Avg.EE and A100 results of CMOF~\cite{msof}, COF~\cite{cof} and \emph{vnflow} variations: \emph{Detail-Aware Weight} (\textbf{DA}) and the fixed weights (\textbf{0}, \textbf{0.5} and \textbf{1}).}\vspace{-4mm}
\label{fig:vnShotShort}
\end{figure*}

\begin{figure}[t!]
\centerline{
\includegraphics[width=0.97\linewidth]{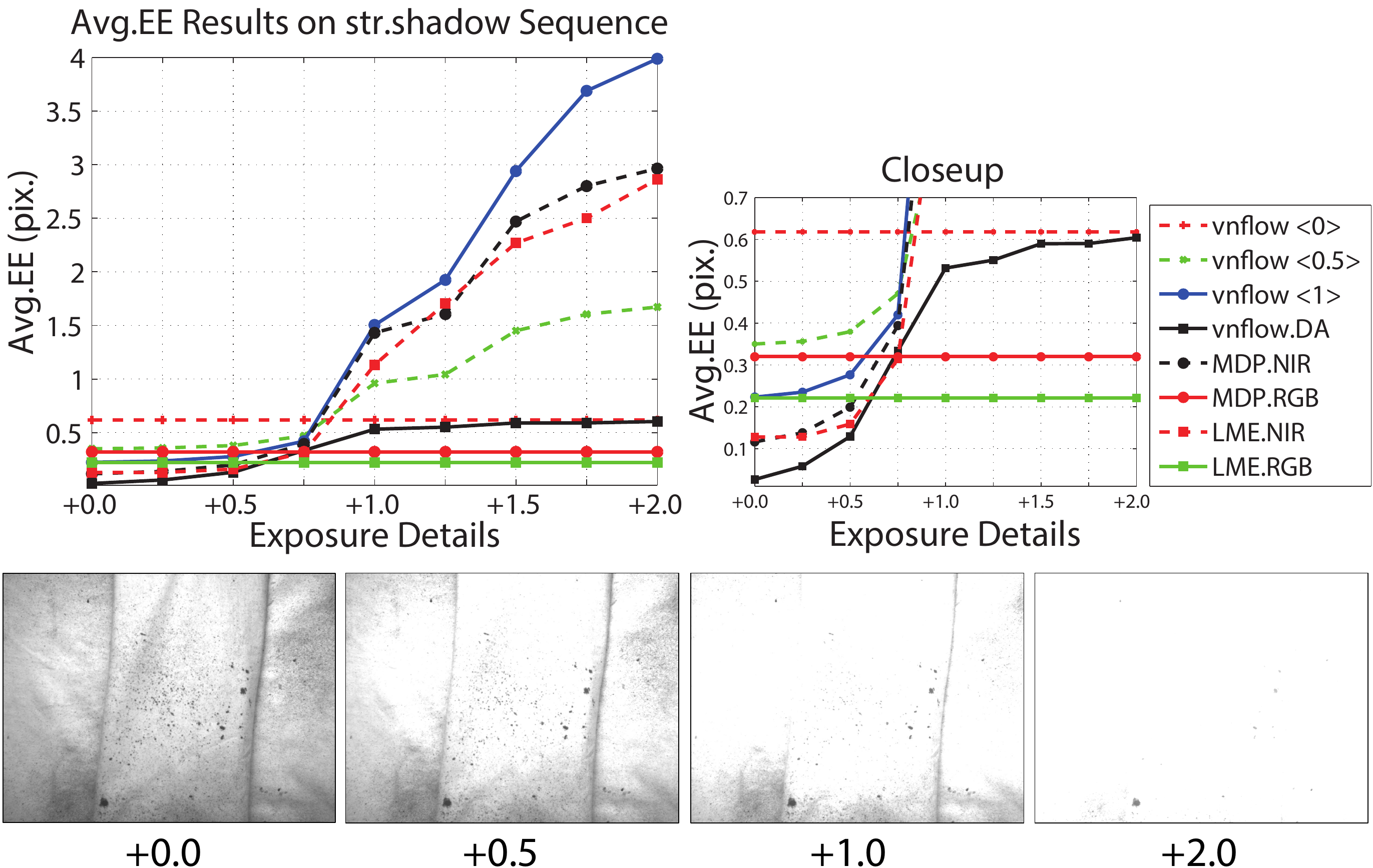}\vspace{-1mm}
}
\caption{Avg.EE measures for \emph{vnflow} on \emph{str.shadow} sequence by varying the exposure (feature distribution) in the NIR channel. \vspace{-1mm}}
\label{fig:expo}
\end{figure}

We consider ten baseline methods in our experiments. Eight of those is executed on the RGB channel of our dataset. The remaining two (MCOF and COF) are evaluated using the NIR channel (and invisible NIR dye GT). Algorithms from Xu~\emph{et al.} (MDP)~\cite{MDP}  (AEE rank 4/119) and Li~\emph{et al.} (LME)~\cite{LME} (rank 11) are state-of-the-art optical flow methods. The former has leading performance in the \emph{Middlebury} evaluation while the latter achieves the state-of-the-art results on Garg~\emph{et al.}~\cite{Garg}. \emph{Combined local-global Optical Flow} (CLG-TV)~\cite{CLG-TV1} (AIE rank 10/119) highlights the utility of bilateral filtering and anisotropic regularization, which gives high performance in image interpolation. \emph{Large Displacement Optical Flow} (LDOF)~\cite{LDOF} (AEE rank 89) is a variational model integrating rich feature descriptors and is designed to overcome large displacement issues. Classic+NL~\cite{Sun10} (rank 28) improves the TV-L1 framework by combining a Lorentzian penalty and a median filtering heuristic. \emph{Horn and Schunck} (HS)~\cite{HS} (rank 108), \emph{Black and Anandan} (BA)~\cite{BA} (rank 101) and \emph{Improved TV-L1} (ITV-L1)~\cite{ITV-L1} (rank 56) are classic modelswidely used in real-world image registration. MCOF~\cite{msof} is considered as the classic approach using both RGB and NIR channels while COF is a robust implementation of~\cite{cof} using additional smoothness constraint~\cite{HS} and coarse-to-fine optimization~\cite{Brox}. Those selected baselines may not cover all the state-of-the-art methods of the community but are able to represent strength/performance in all typical measures.

We first perform an evaluation on the short sequences of our GT dataset (i.e. each of the above algorithms are executed on the RGB channel only). Fig.~\ref{fig:sceneShotShort} shows a screen shot of our public evaluation website where eight optical flow methods are quantitatively compared to each other using their default parameter settings. Note that the relative \emph{Middlebury} AEE rank (Average rank, captured on Feb. 23, 2016) of the baseline methods is also listed for comparison. We observe that LME leads all trials in Avg.EE. ITV-L1 and Classic-NL respectively rank 2.50 and 3.25 in general Avg.EE. The former outperforms most other algorithms in \emph{featureless} while the latter shows more robust toward flow discontinuities (\emph{mObjs}, \emph{triObjs} and \emph{crease}) and blur motion (\emph{blur}). Note that most methods have a large error (\textgreater 0.5 Avg.EE.) for \emph{illumination} because the strong illumination change violates the \emph{Intensity Consistency}. In this case, LME (Avg.EE 0.09), ITV-L1 (Avg.EE 0.11) and LDOF (Avg.EE 0.29) give higher performance over the other methods.

Interestingly, compared to \emph{Middlebury} the short sequences of our dataset result in a significantly different ranking. We believe this is due to the range of new photometric effects in our GT which are absent in \emph{Middlebury}. MDP achieves top performance in \emph{Middlebury} but ranks (in relative terms) 6 in \emph{featureless} and 4.13 in Avg.EE by average. This is because large textureless regions in \emph{featureless} provide less SIFT features, in turn weakening its inner motion detail preservation. Additionally, LME ranks higher (in relative terms) than in \emph{Middllebury}. The reason may be due to the local smoothness and deformation penalties~\cite{LME}, which is robust to complex motion (Avg.EE 0.12 in \emph{blur}) and textureless regions (Avg.EE 0.09 in \emph{featureless}).

\begin{figure}[t!]
\centerline{
\includegraphics[width=1\linewidth]{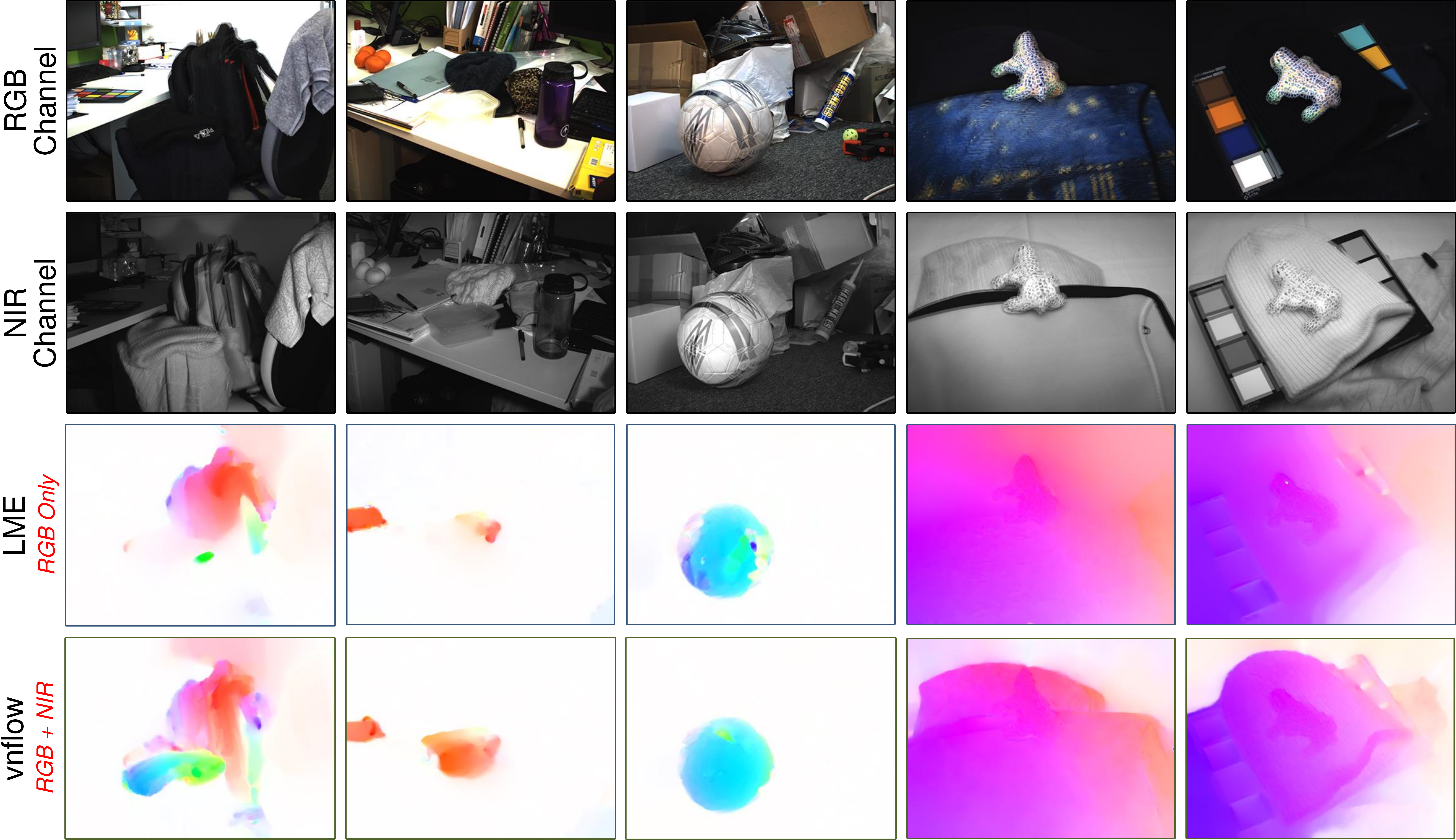}\vspace{-1mm}
}
\caption{Visual results of \emph{vnflow} on real-world sequences (\textbf{From Left To Right}) \emph{hat}, \emph{office}, \emph{football}, \emph{arts} and \emph{dark}. Note that we show the overlaps of two input images in the top two rows. Hence the blur is not caused by the image quality. \vspace{-5mm}}
\label{fig:realworld}
\end{figure}

An evaluation on the RGB channel of the long sequences is also performed as shown in Fig.~\ref{fig:sceneShotLong}. Similar to the short sequence case, LME provides the best Avg.EE in all trials while Classic+NL, ITV-L1 and MDP yield equally top performance in \emph{stretch}. All the methods display comparatively larger Avg.EE in \emph{mBlur} due to the camera blur and fast motion in the scene. In the robustness test (SD), ITV-L1 reaches the top performance on both \emph{crush} and \emph{stretch} while LME yields the best results on the other sequences. Our graph view in Fig.~\ref{fig:sceneShotLong}(b) shows that both LME and ITV-L1 give lower accumulated error (Acc.EE) against the other baselines along the entire \emph{crush} sequence.

To evaluate our hybrid RGB-NIR optical flow algorithm -- and the potential benefit of using our weighting scheme and multiple spectrums for dense tracking -- we compare our method which includes the proposed \emph{Detail-Aware Weight} (\emph{vnflow.DA}) against MCOF, COF as well as three other implementations using fixed weights (0, 0.5 and 1) in Fig.~\ref{fig:vnShotShort}. Note that our implementation of COF is applied using all R, G, B and NIR channels. It is observed that \emph{vnflow.DA} outperforms all other baseline methods in Avg.EE in all cases. Our algorithm \textbf{without} NIR energy ($\lambda = 0$) shows low overall performance (Avg.EE rank 7.75) while \textbf{with only} NIR energy ($\lambda = 1$) it ranks 4.38 in Avg.EE. In addition, LME with NIR imagery achieves comparably lower overall Avg.EE but shows large A100 error in \emph{str.shadow} due to the large shadow that affects the inner detail preservation process. MCOF takes the advantage from additional NIR channel and gives precision generally closed to our methods using fixed weights. In some difficult cases e.g. \emph{mObjs} etc, the precision of MCOF is affected by the primitive optimization scheme. Furthermore, COF yields competitive performance in overall (Avg.EE rank 4.5) and shows the improvement over the case used RGB channel only (Avg.EE rank 7.75).

We then perform an Avg.EE comparison of LME, MDP and four \emph{vnflow} implementations on \emph{str.shadow} by varying the feature distribution in the NIR channel. As shown in Fig.~\ref{fig:expo}, we are ramping up the exposure to reduce the overall number of NIR features in the image. As expected, less NIR information (higher exposure) generally increases the Avg.EE. However, even with a very low quantity of NIR information (+2.0), \emph{vnflow.DA} still shows improvement over other implementations using fixed weights (0, 0.5 and 1).


Fig.~\ref{fig:realworld}, a compelling illustration, visualises how switching between RGB and NIR information can contribute to the strong performance of \emph{vnflow.DA}. Our \emph{vnflow.DA} uses texture details invisible in the RGB channel where required (and vice-versa). This provides an explanation to why the algorithm gives higher accuracy against other methods which are using either the RGB or NIR channels alone. However, it should be noted that our evaluation here is a relative one -- providing the first insight into how optical flow (and other tracking) can potentially benefit from multiple spectrums. An absolute RGB-NIR evaluation would require a \emph{third} hidden spectrum -- in the same way that to evaluate RGB algorithms in our new dataset and evaluation framework we have required NIR for GT (i.e. a second spectrum). Such an evaluation of dedicated RGB/NIR (or other multispectral) methods may not be practical until multispectral tracking, hardware and other suitable dyes become more widespread.


\section{Conclusion}


In this paper, we present a new publicly available ground truth dataset for evaluating RGB/Color based optical flow algorithms. By leveraging RGB-NIR imaging and NIR visible dyes, our dataset provides dense ground truth for real-world objects in short and long sequences, as well as with nonrigid motion, illumination changes and motion blur. Algorithms are executed on the RGB sequences, and their result is compared to the ground truth obtained by analysing the dense patters only visible in the NIR channel. We also propose a multispectral optical flow framework which utilizes an adoptive weighting scheme to balance the contributions of different channels in order to optimize overall performance. This provides a compelling insight into the potential benefits for tracking in multiple spectra. One further challenge is finding a dye solution which remains invisible in the RGB channel for any object surface. This way, ground truth deformations could be obtained from a wider range of material.

{\small
\bibliographystyle{IEEEtran}
\bibliography{egbib}
}

\end{document}